\documentclass[11pt,a4paper]{article}
\usepackage[hyperref]{acl2019}
\usepackage{times}
\usepackage{latexsym}
\usepackage{url}
\usepackage{enumerate}
\usepackage{graphicx}
\usepackage{booktabs}
\usepackage{colortbl}
\usepackage[german]{babel}
\usepackage{csquotes}
\usepackage{subcaption}

\usepackage{tablefootnote}
\usepackage{longtable}

\usepackage{url, amssymb, amsmath, amsthm, enumitem}

\aclfinalcopy

\newcommand{\fairseq}{\textsc{fairseq}}

\newcommand{\entode}{En$\rightarrow$De}
\newcommand{\detoen}{De$\rightarrow$En}

\newcommand{\entoru}{En$\rightarrow$Ru}
\newcommand{\rutoen}{Ru$\rightarrow$En}

\newcommand{\enbide}{En$\leftrightarrow$De}
\newcommand{\enbiru}{En$\leftrightarrow$Ru}

\def\Const{\rowcolor{white}}
\def\UnCon{\rowcolor{lightgray}}

\title{Facebook FAIR's WMT19 News Translation Task Submission}

\author{Nathan Ng, Kyra Yee, Alexei Baevski, Myle Ott, Michael Auli, Sergey Edunov \\
Facebook AI Research, \\Menlo Park, CA \& New York, NY.}

\date{}

\begin{document}
\maketitle
\begin{abstract}
This paper describes Facebook FAIR's submission to the WMT19 shared news translation task.
We participate in two language pairs and four language directions, English $\leftrightarrow$ German and English $\leftrightarrow$ Russian.
Following our submission from last year, our baseline systems are large BPE-based transformer models trained with the \fairseq{} sequence modeling toolkit which rely on sampled back-translations.
This year we experiment with different bitext data filtering schemes, as well as with adding filtered back-translated data.
We also ensemble and fine-tune our models on domain-specific data, then decode using noisy channel model reranking.
Our submissions are ranked first in all four directions of the human evaluation campaign. 
On \entode{}, our system significantly outperforms other systems as well as human translations. 
This system improves upon our WMT'18 submission by 4.5 BLEU points.
\end{abstract}

\section{Introduction}
We participate in the WMT19 shared news translation task in two language pairs and four language directions, English$\rightarrow$German (\entode{}), German$\rightarrow$English (\detoen{}), English$\rightarrow$Russian (\entoru{}), and Russian$\rightarrow$English (\rutoen{}). 
Our methods are based on techniques and approaches used in our submission from last year \cite{edunov2018understanding}, including the use of subword models, \cite{sennrich2016neural}, large-scale back-translation, and model ensembling.
We train all models using the \fairseq{} sequence modeling toolkit \cite{ott2019fairseq}. 
Although document level context for \entode{} is now available,  all our systems are pure sentence level systems.
In the future, we expect better results from leveraging this additional context information.

Compared to our WMT18 submission, we also decide to compete in the \enbiru{} and \detoen{} translation directions. 
Although all four directions are considered high resource settings where large amounts of bitext data is available, we demonstrate that leveraging high quality monolingual data through back-translation is still very important. 
For all language directions, we back-translate the Newscrawl dataset using a reverse direction bitext system.
In addition to back-translating the relatively clean Newscrawl dataset, we also experiment with back-translating portions of the much larger and noisier Commoncrawl dataset.
For our final models, we apply a domain-specific fine-tuning process and decode using noisy channel model reranking (Anonymous, 2019).

Compared to our WMT18 submission in the \entode{} direction, we observe substantial improvements of 4.5 BLEU.
Some of these gains can be attributed to differences in dataset quality, but we believe most of the improvement comes from larger models, larger scale back-translation, and noisy channel model reranking with strong channel and language models.

\section{Data}
For the \enbide{} language pair we use all available bitext data  including the bicleaner version of Paracrawl. 
For our monolingual data we use English and German Newscrawl. 
Although our language models were trained on document level data, we did not use document level boundaries in our final decoding step, so all our systems are purely sentence level systems.

For the \enbiru{} language pair we also use all available bitext data. 
For our monolingual data we use English and Russian Newscrawl as well as a filtered portion of Russian Commoncrawl.
We choose to use Russian Commoncrawl to augment our monolingual data due to the relatively small size of Russian Newscrawl compared to English and German.  

\subsection{Data Preprocessing}
Similar to last year's submission for \entode{}, we normalize punctuation and tokenize all data with the Moses tokenizer \cite{koehn:moses:2007}. 
For \enbide{} we use joint byte pair encodings (BPE) with 32K split operations for subword segmentation \cite{sennrich2016neural}. 
For \enbiru{}, we learn separate BPE encodings with 24K split operations for each language. 
Systems trained with this separate BPE encoding performed significantly better than those trained with joint BPE. 

\subsection{Data Filtering}
\subsubsection{Bitext}
Large datasets crawled from the internet are naturally very noisy and can potentially decrease the performance of a system if they are used in their raw form.
Cleaning these datasets is an important step to achieving good performance on any downstream tasks.

We apply language identification filtering (\texttt{langid}; Lui et al., 2012)\nocite{lui2012langid}, keeping only sentence pairs with correct languages on both sides. 
Although not the most accurate method of language identification \cite{joulin2016bag}, one side effect of using \texttt{langid} is the removal of very noisy sentences consisting of mostly garbage tokens, which are classified incorrectly and filtered out. 

We also remove sentences longer than 250 tokens  as  well  as  sentence pairs with  a  source/target  length  ratio  exceeding  1.5. 
In total, we filter out about 30\% of the original bitext data.
See Table~\ref{tab:bitext_size} for details on the bitext dataset sizes.

\subsubsection{Monolingual}
For monolingual Newscrawl data we also apply \texttt{langid} filtering. 
Since the monolingual Newscrawl corpus for Russian is significantly smaller than that of German or English, we augment our monolingual Russian data with data from the commoncrawl corpus.
Commoncrawl is the largest monolingual corpus available for training but is also very noisy.
In order to select a limited amount of high quality, in-domain sentences from the larger corpus, we adopt the method of~\citet{moore2010intelligent} for selecting in-domain data (\textsection\ref{subsection:btcc}).

\begin{table}[t]
\centering
\begin{tabular}{lcc}
\toprule
& En-De & En-Ru \\ \midrule
No filter & 38.8M & 38.5M \\
+ length filter & 35.7M & 33.4M \\
+ \texttt{langid} filter & 27.7M & 26.0M \\
\bottomrule
\end{tabular}
\caption{Number of sentences in bitext datasets for different filtering schemes}

\label{tab:bitext_size}
\end{table}

\section{System Overview}
\setquotestyle[quotes]{german}
\setlength{\abovedisplayskip}{5pt}
\setlength{\belowdisplayskip}{5pt}

\subsection{Base System}
Our base system is based on the big Transformer architecture \cite{vaswani2017attention} as implemented in \fairseq{}. 
We experiment with increasing network capacity by increasing embed dimension, FFN size, number of heads, and number of layers.
We find that using a larger FFN size (8192) gives a reasonable improvement in performance while maintaining a manageable network size. 
All subsequent models, including ensembles, use this larger FFN Transformer architecture.

We trained all our models using \fairseq{}~\cite{ott2019fairseq} on 128 Volta GPUs, following the setup described in~\citet{ott:scaling:2018}

\subsection{Large-scale Back-translation}
Back-translation is an effective and commonly used data augmentation technique to incorporate monolingual data into a translation system. 
Back-translation first trains an intermediate target-to-source system that is used to translate monolingual target data into additional synthetic parallel data.  
This data is used in conjunction with human translated bitext data to train the desired source-to-target system. 

In this work we used back-translations obtained by sampling~\cite{edunov2018understanding} from an ensemble of three target-to-source models. 
We found that models trained on data back-translated using an ensemble instead of a single model performed better (Table~\ref{tab:bt_ensemble_vs_single}). 
Previous work also found that upsampling the bitext data can improve back-translation~\citep{edunov2018understanding}.
We adopt this method to tune the amount of bitext and synthetic data the model is trained on.
We find a ratio of 1:1 synthetic to bitext data to perform the best.

\begin{table}[t]
\centering
\begin{tabular}{lcc}
\toprule
& \multicolumn{2}{c}{\entoru{}} \\
& Single Model & Ensemble \\ \midrule
newstest15 & 35.98 & 36.32 \\
newstest16 & 32.78 & 33.28 \\
newstest17 & 36.57 & 36.77 \\
newstest18 & 34.72 & 34.72 \\
\bottomrule
\end{tabular}
\caption{SacreBLEU for English-Russian models trained with data back-translated using a single model vs. an ensemble of two models}
\label{tab:bt_ensemble_vs_single}
\end{table}

\begin{table}[t]
\centering
\begin{tabular}{lccc}
\toprule
& En & De & Ru \\ \midrule
Newscrawl & 434M & 559M & 80M \\
+ \texttt{langid} filter & 424M & 521M & 76M\\
\midrule
Commoncrawl & - & - & 1.2B \\
+ \texttt{KenLM} filter & - & - & 60M \\
\midrule
\textbf{Total} & 424M & 521M & 136M \\
\bottomrule
\end{tabular}
\caption{Number of sentences in monolingual datasets available for back-translation}
\label{tab:monolingual_size}
\end{table}

\subsubsection{Back-translating Commoncrawl}
\label{subsection:btcc}
The amount of monolingual Russian data available in the Newscrawl dataset is significantly smaller than that of English and German (Table \ref{tab:monolingual_size}).
In order to increase the amount of monolingual Russian data for back-translation, we experiment with incorporating Commoncrawl data.
Commoncrawl is a much larger and noisier dataset compared to Newscrawl, and is also non-domain specific.
We experiment with methods to identify a subset of Commoncrawl that is most similar to Newscrawl.
Specifically, we use the in-domain filtering method described in~\citet{moore2010intelligent}.

Given an in domain corpus $I$, in this case Newscrawl, and a non-domain specific corpus $N$, in this case Commoncrawl, we would like the find the subcorpus $N_I$ that is drawn from the same distribution as $I$.
For any given sentence $s$, we can calculate, using Bayes' rule, the probability a sentence $s$ in $N$ is drawn from $N_I$
\begin{equation}
    P(N_I | s, N) = \frac{P(s|N_I)P(N_I|N)}{P(s|N)
}\end{equation}
We ignore the $P(N_I|N)$ term, since it will be constant for any given $I$ and $N$, and use $P(s|I)$ instead of $P(s|N_I)$, since $I$ and $N_I$ are drawn from the same distribution.
Moving into the log domain, we can calculate the probability score for a sentence $s$ by $\log P(N_I | s, N) = \log P(s|I) - \log P(s|N)$, or after normalizing for length, $H_I(s) - H_N(s)$, where $H_I(s)$ and $H_N(s)$ are the word-normalized cross entropy scores for a sentence $s$ according to language models $L_I$ and $L_N$ trained on $I$ and $N$ respectively.

Our corpora are very large and we therefore use an $n$-gram model~\citep{kenneth2011kenlm} rather than a neural language model which would be much slower to train and evaluate.
We train two language models $L_I$ and $L_N$ on Newscrawl and Commoncrawl respectively, then score every sentence $s$ in Commoncrawl by $H_I(s) - H_N(s)$.
We select a cutoff of $0.01$, and use all sentences that score higher than this value for back-translation, or about 5\% of the entire dataset. 

\subsection{Fine-tuning}
Fine-tuning with domain-specific data is a common and effective method to improve translation quality for a downstream task. 
After completing training on the bitext and back-translated data, we train for an additional epoch on a smaller in-domain corpus.
For \detoen{}, we fine-tune on test sets from previous years, including newstest2012, newstest2013, newstest2015, and newstest2017.
For \entode{}, we fine-tune on previous test sets as well as the News-Commentary dataset.
For \enbiru{} we fine-tune on a combination of News-Commentary, newstest2013, newstest2015, and newstest2017. 
The other test sets are held out for other tuning procedures and evaluation metrics.

\subsection{Noisy Channel Model Reranking}

\begin{table}[t]
\centering
\begin{tabular}{lcccc}
\toprule
& En-\textbf{De} & De-\textbf{En} & En-\textbf{Ru} & Ru-\textbf{En} \\ \midrule
newstest12 & 26.7 & 28.0 & - & - \\
newstest13 & 27.8 & 27.6 & 42.7 & 27.6 \\
newstest14 & 21.4 & 24.0 & 32.3 & 22.4 \\
newstest15 & 25.1 & 24.6 & 34.7 & 21.8 \\
newstest16 & 24.5 & 22.0 & 35.5 & 19.4 \\
newstest17 & 25.0 & 21.9 & 37.9 & 19.5 \\
newstest18 & 25.1 & 26.0 & 39.3 & 20.0 \\
\bottomrule
\end{tabular}
\caption{Perplexity scores for language models on bolded target languages in all translation directions}
\label{tab:lm_newstest}
\end{table}

$N$-best reranking is a method of improving translation quality by scoring and selecting a candidate hypothesis from a list of $n$-best hypotheses generated by a source-to-target, or forward model.
For our submissions, we rerank using a noisy channel model approach.

Given a target sequence $y$ and a source sequence $x$, the noisy channel approach applies Bayes' rule to model 
\begin{equation}
    P(y|x) = \frac{P(x|y)P(y)}{P(x)}
\end{equation}
Since $P(x)$ is constant for a given source sequence $x$, we can ignore it.
We refer to the remaining terms $P(y|x)$, $P(x|y)$, and $P(y)$, as the forward model, channel model, and language model respectively. 
In order to combine these scores for reranking, we calculate for every one of our $n$-best hypotheses:
\begin{equation}
    \log P(y|x) + \lambda_1 \log P(x|y) + \lambda_2 \log P(y) 
\end{equation} 
The weights $\lambda_1$ and $\lambda_2$ are determined by tuning them with a random search on a validation set and selecting the weights that give the best performance.
In addition, we also tune a length penalty.

For all translation directions, our forward models are ensembles of fine-tuned and back-translated models.
Since we compete in both directions for both language pairs, for any given translation direction we can use the forward model for the reverse direction as the channel model.
Our language models for each of the target languages English, German, and Russian, are big Transformer decoder models with FFN 8192.
We train the language models on the monolingual Newscrawl dataset, and use document level context for the English and German models. 
Perplexity scores for the language models on the bolded target language of each translation direction are shown in table \ref{tab:lm_newstest}.
With a smaller amount of monolingual Russian data available, we observe that our Russian language model performs worse than the German and English language models.

To select the length penalty and weights, $\lambda_1$ and $\lambda_2$, for decoding, we use random search, choosing values in the range $[0, 2)$ for the weights and values in the range $[0, 1)$ for the length penalty.
For all language directions, we choose the weights that give the highest BLEU score on a combined dataset of newstest2014 and newstest2016.

To run our final decoding step, we first use the forward model with beam size $50$ to generate an $n$-best list.
We then use the channel and language models to score each of these hypotheses, using the weights and length penalty tuned previously.
Finally, we select the hypothesis with the highest score as our output.

\subsection{Postprocessing}
\setquotestyle[quotes]{german}
For \entode{} and \entoru{}, we also change the standard English quotation marks (`` ... '') to German-style quotation marks (\enquote{ ... }).

\section{Results}
Results and ablations for \entode{} are shown in Table~\ref{tab:ablation_ende}, \detoen{} in Table~\ref{tab:ablation_deen}, \entoru{} in Table~\ref{tab:ablation_enru} and \rutoen{} in Table~\ref{tab:ablation_ruen}.
We report case-sensitive SacreBLEU scores using SacreBLEU~\cite{matt2018clarity}\footnote{SacreBLEU signatures:\newline{} BLEU+case.mixed+lang.en-de+numrefs.1+smooth.exp+\newline{}test.wmt\{17/18\}+tok.13a+version.1.2.11, BLEU+case.mixed+lang.de-en+numrefs.1+smooth.exp+\newline{}test.wmt\{17/18\}+tok.13a+version.1.2.11, BLEU+case.mixed+lang.ru-en+numrefs.1+smooth.exp+\newline{}test.wmt\{17/18\}+tok.13a+version.1.2.11, BLEU+case.mixed+lang.en-ru+numrefs.1+smooth.exp+\newline{}test.wmt\{17/18\}+tok.intl+version.1.2.11}, using international tokenization for \entoru{}.
In the final row of each table we also report the case-sensitive BLEU score of our submitted system on this year's test set.
All single models and individual models within ensembles are averages of the last $10$ checkpoints of training.
Our baseline systems are big Transformers as described in~\cite{vaswani2017attention}.
The baselines were trained with minimally filtered data, removing only those sentences longer than 250 words and exceeding a source/target length ratio of $1.5$
This setup gave us a reasonable baseline to evaluate data filtering.

\subsection{English$\rightarrow$German}
\begin{table}[t]
\centering
\begin{tabular}{lcc}
\toprule
& \multicolumn{2}{c}{\textbf{\entode{}}} \\
System & \bf news2017 & \bf news2018 \\ \midrule
baseline & 30.90 & 45.40 \\ 
+ langid filtering & 30.78 & 46.43 \\
+ ffn 8192 & 31.15 & 46.28 \\
+ BT & 33.62 & 46.66 \\
+ fine tuning & - & 47.61\\
+ ensemble & - & 49.27 \\
+ reranking & - & 50.63 \\
WMT'18 submission & - & 46.10 \\
\midrule
\bf WMT'19 submission & \multicolumn{2}{c}{\bf 42.7} \\ 
\bottomrule
\end{tabular}
\caption{SacreBLEU scores on English$\rightarrow$German. 
}
\label{tab:ablation_ende}
\end{table}

For \entode{}, \texttt{langid} filtering, larger FFN, and ensembling improve our baseline performance on news2018 by about 1.5 BLEU.
Note that our best bitext only systems already outperforms our system from last year by 1 BLEU point.
This is perhaps due to the addition of higher quality bitext data and improved data filtering techniques.
The addition of back-translated (BT) data improves single model performance by only 0.3 BLEU, but combining this with fine-tuning and ensembling gives us a total of 3 BLEU.
Finally, applying reranking on top of these strong ensembled systems gives another 1.4 BLEU.

\subsection{German$\rightarrow$English}
\begin{table}[t]
\centering
\begin{tabular}{lcc}
\toprule
& \multicolumn{2}{c}{\textbf{\detoen{}}} \\
System & \bf news2017 & \bf news2018 \\ \midrule
baseline & 37.28  & 45.32 \\ 
+ langid and ffn 8192 & 38.45 & 46.16 \\
+ BT & 41.08 & 48.78 \\
+ fine tuning & - & 49.07 \\
+ ensemble & - & 49.60 \\
+ reranking & - & 51.13\\
\midrule
\bf WMT'19 submission & \multicolumn{2}{c}{\bf 40.8} \\ 
\bottomrule
\end{tabular}
\caption{SacreBLEU scores on German$\rightarrow$English.}
\label{tab:ablation_deen}
\end{table}

For \detoen{}, as with \entode{}, we see similar improvements with \texttt{langid} filtering, larger FFN, and ensembling on the order of 1.4 BLEU.
Compared to \entode{} however, we also observe that the addition of back-translated data is much more significant, improving single model performance by over 2.5 BLEU. 
Fine-tuning, ensembling, and reranking add an additional 2.4 BLEU, with reranking contributing 1.5 BLEU, a majority of the improvement.

\subsection{English$\rightarrow$Russian}
\begin{table}[t]
\centering
\begin{tabular}{lcc}
\toprule
& \multicolumn{2}{c}{\textbf{\entoru{}}} \\
System & \bf news2017 & \bf news2018 \\ \midrule
baseline & 35.42 & 31.53 \\ 
+ langid filtering & 35.69 & 31.77 \\
+ ffn 8192 & 36.66 &  33.49\\
+ BT NewsCrawl & 40.09 &  37.07 \\
+ BT CommonCrawl & 40.42 & 37.3 \\
+ fine tuning & - & 37.74 \\
+ ensemble & -  & 38.59 \\
+ reranking & -  & 39.53 \\
\midrule
\bf WMT'19 submission & \multicolumn{2}{c}{\bf 36.3} \\ 
\bottomrule
\end{tabular}
\caption{SacreBLEU scores on English$\rightarrow$Russian }
\label{tab:ablation_enru}
\end{table}

For \entoru{}, we observe large improvements of 2.4 BLEU over a bitext-only model after applying \texttt{langid} filtering, larger FFN, and ensembling. 
Since we start with a lower quality initial \enbiru{} bitext dataset, we observe a large improvement of 3.5 BLEU by adding back-translated data.
Augmenting this back-translated data with Commoncrawl adds an additional 0.2 BLEU.
Finally, applying fine-tuning, ensembling, and reranking adds 2.2 BLEU, with reranking contributing 1 BLEU.

\subsection{Russian$\rightarrow$English}
\begin{table}[t]
\centering
\begin{tabular}{lcc}
\toprule
& \multicolumn{2}{c}{\textbf{\rutoen{}}} \\
System & \bf news2017 & \bf news2018 \\ \midrule
baseline & 37.07 & 32.69 \\ 
+ langid and ffn 8192 & 37.72 & 33.44 \\
+ BT & 41.68 & 36.49 \\
+ fine tuning & - & 38.54 \\
+ ensemble & - & 38.96 \\
+ reranking & -  & 40.16 \\
\midrule
\bf WMT'19 submission & \multicolumn{2}{c}{\bf 40.0} \\ 
\bottomrule
\end{tabular}
\caption{SacreBLEU scores on Russian$\rightarrow$English }
\label{tab:ablation_ruen}
\end{table}

For \rutoen{}, we observe similar trends to \enbide{}, with \texttt{langid} filtering, larger FFN, and ensembling improving performance of a bitext-only system by 1.6 BLEU.
Backtranslation adds 3 BLEU, again most likely due to the lower quality bitext data available.
Fine-tuning, ensembling, and reranking add almost 4 BLEU, with reranking contributing 1.2 BLEU.

\subsection{Reranking}
For every language direction, reranking gives a significant improvement, even when applied on top of an ensemble of very strong back-translated models. 
We also observe that the biggest improvement of 1.5 BLEU comes in the \detoen{} language direction, and the smallest improvement of 1 BLEU in the \entoru{} direction.
This is perhaps due to the relatively weak Russian language model, which is trained on significantly less data compared to English and German.
Improving our language models may lead to even greater improvements with reranking.

\subsection{Human Evaluations}

All our systems participated in the human evaluation campaign of WMT'19.
For different systems, different styles of evaluations were used. 
All our systems except \rutoen{} were evaluated with document level context and had a document level rating collected.
Source based direct assessment was used for systems translating from English, and target based direct assessment was used for systems translating to English.
See Table~\ref{tab:types_of_human_eval} for more details. 

\begin{table}[t]
\small
\centering
\begin{tabular}{lccc}
\toprule
         & Doc Rating $+$ & Seg Rating $+$        & Seg Rating $-$\\
         & Doc Context    & Doc Context           & Doc Context \\

         & (DR$+$DC) & (SR$+$DC) & (SR$-$DC) \\ 
\midrule
de-en & M & {\bf M}  &   \\
en-de & B & {\bf B}  &   \\
en-ru & B & {\bf B}  &   \\
ru-en &  &  & {\bf M}  \\
\bottomrule
\end{tabular}
\caption{Human evaluation configurations;
M denotes monolingual human evaluation, or target-based direct assessment, where translations are compared to human references; 
B denotes bilingual/source based evaluation where the human annotators evaluate MT
output based only on the source sentence (and no reference translation is present); 
+DC denotes systems evaluated with document level context, 
-DC without document context.}
\label{tab:types_of_human_eval}
\end{table}

Facebook-FAIR was ranked first in all four language directions we compete in. 
Table~\ref{tab:ende_human} shows that our \entode{} submission significantly outperforms other systems as well as human translations. 
Our submissions for \detoen{}, \entoru{} and \rutoen{} also achieve the highest score. 

Although our systems are pure sentence-level models, they performed well irrespective of whether the evaluation method used document context or not. 
For document level rankings, our \entode{} system also ranked first and significantly outperformed human translations. 
Our \entoru{} submission achieved the highest score among all submissions and is tied for the first place with human translations. 
The \detoen{} system achieved the second highest score among constrained systems. See \citep{bojar-wmt19} for details.

\begin{table}[t]
\small
\centering
\begin{tabular}{ccrl}
 & Ave. & Ave. z & System\\ \hline
\Const{} & 90.3 & 0.347 & Facebook-FAIR \\ \hline
\Const{} & 93.0 & 0.311 & Microsoft-WMT19-sent-doc \\
\Const{} & 92.6 & 0.296 & Microsoft-WMT19-doc-level \\
\UnCon{} & 90.3 & 0.240 & HUMAN \\
\Const{} & 87.6 & 0.214 & MSRA-MADL \\
\Const{} & 88.7 & 0.213 & UCAM \\
\Const{} & 89.6 & 0.208 & NEU \\
\Const{} & 87.5 & 0.189 & MLLP-UPV \\
\Const{} & 87.5 & 0.130 & eTranslation \\
\Const{} & 86.8 & 0.119 & dfki-nmt \\
\UnCon{} & 84.2 & 0.094 & online-B \\
\Const{} & 86.6 & 0.094 & Microsoft-WMT19-sent-level \\
\Const{} & 87.3 & 0.081 & JHU \\
\Const{} & 84.4 & 0.077 & Helsinki-NLP \\
\UnCon{} & 84.2 & 0.038 & online-Y \\
\Const{} & 83.7 & 0.010 & lmu-ctx-tf-single \\
\UnCon{} & 84.1 & 0.001 & PROMT-NMT \\
\UnCon{} & 82.8 & $-$0.072 & online-A \\
\UnCon{} & 82.7 & $-$0.119 & online-G \\
\Const{} & 80.3 & $-$0.129 & UdS-DFKI \\
\Const{} & 82.4 & $-$0.132 & TartuNLP-c \\ \hline
\UnCon{} & 76.3 & $-$0.400 & online-X \\ \hline
\UnCon{} & 43.3 & $-$1.769 & en-de-task \\
\end{tabular}
\caption{Official results of the WMT'19 \entode{} News Translation Task. Systems are ordered by DA z-score; systems within a cluster are considered tied; grayed entries indicate systems using resources beyond the provided data.}
\label{tab:ende_human}
\end{table}

\section{Conclusions}
This paper describes Facebook FAIR's submission to the WMT19 news translation task.
For all four translation directions, \enbide{} and \enbiru{}, we use the same strategy of filtering bitext data, performing sampling-based back-translation on monolingual data, then training strong individual models on a combination of this data.
Each of these models is fine-tuned and ensembled into a final system that is used for decoding with noisy channel model reranking.
We demonstrate the effectiveness of our noisy channel-based reranking approach even when applied on top of very strong systems, and rank first in all four directions of the human evaluation campaign.

\bibliography{acl2019}

\begin{thebibliography}{12}
\expandafter\ifx\csname natexlab\endcsname\relax\def\natexlab#1{#1}\fi

\bibitem[{Bojar et~al.(2019)Bojar, Federmann, Fishel, Graham, Haddow, Huck,
  Koehn, and Monz}]{bojar-wmt19}
Ond\v{r}ej Bojar, Christian Federmann, Mark Fishel, Yvette Graham, Barry
  Haddow, Matthias Huck, Philipp Koehn, and Christof Monz. 2019.
\newblock Findings of the 2019 conference on machine translation (wmt19).
\newblock In \emph{Proceedings of the Fourth Conference on Machine Translation,
  Volume 2: Shared Task Papers}, Florence, Italy. Association for Computational
  Linguistics.

\bibitem[{Edunov et~al.(2018)Edunov, Ott, Auli, and
  Grangier}]{edunov2018understanding}
Sergey Edunov, Myle Ott, Michael Auli, and David Grangier. 2018.
\newblock Understanding back-translation at scale.
\newblock In \emph{Proceedings of the 2018 Conference on Empirical Methods in
  Natural Language Processing}, pages 489--500.

\bibitem[{Heafield(2011)}]{kenneth2011kenlm}
Kenneth Heafield. 2011.
\newblock Kenlm: faster and smaller language model queries.
\newblock In \emph{Proceedings of the Sixth Workshop on Statistical Machine
  Translation}, pages 187--197.

\bibitem[{Joulin et~al.(2016)Joulin, Grave, Bojanowski, and
  Mikolov}]{joulin2016bag}
Armand Joulin, Edouard Grave, Piotr Bojanowski, and Tomas Mikolov. 2016.
\newblock Bag of tricks for efficient text classification.
\newblock \emph{arXiv preprint arXiv:1607.01759}.

\bibitem[{Koehn et~al.(2007)Koehn, Hoang, Birch, Callison-Burch, Federico,
  Bertoldi, Cowan, Shen, Moran, Zens, Dyer, Bojar, Constantin, and
  Herbst}]{koehn:moses:2007}
Philipp Koehn, Hieu Hoang, Alexandra Birch, Chris Callison-Burch, Marcello
  Federico, Nicola Bertoldi, Brooke Cowan, Wade Shen, Christine Moran, Richard
  Zens, Chris Dyer, Ondrej Bojar, Alexandra Constantin, and Evan Herbst. 2007.
\newblock Moses: Open source toolkit for statistical machine translation.
\newblock In \emph{ACL Demo Session}.

\bibitem[{Lui and Baldwin(2012)}]{lui2012langid}
Marco Lui and Timothy Baldwin. 2012.
\newblock langid. py: An off-the-shelf language identification tool.
\newblock In \emph{Proceedings of the ACL 2012 system demonstrations}, pages
  25--30. Association for Computational Linguistics.

\bibitem[{Moore and Lewis(2010)}]{moore2010intelligent}
Robert Moore and William Lewis. 2010.
\newblock Intelligent selection of language model training data.
\newblock In \emph{Proceedings of the ACL 2010 Conference Short Papers}, pages
  220--224.

\bibitem[{Ott et~al.(2019)Ott, Edunov, Baevski, Fan, Gross, Ng, Grangier, and
  Auli}]{ott2019fairseq}
Myle Ott, Sergey Edunov, Alexei Baevski, Angela Fan, Sam Gross, Nathan Ng,
  David Grangier, and Michael Auli. 2019.
\newblock fairseq: A fast, extensible toolkit for sequence modeling.
\newblock In \emph{Proceedings of NAACL-HLT 2019: Demonstrations}.

\bibitem[{Ott et~al.(2018)Ott, Edunov, Grangier, and Auli}]{ott:scaling:2018}
Myle Ott, Sergey Edunov, David Grangier, and Michael Auli. 2018.
\newblock Scaling neural machine translation.
\newblock In \emph{Proc. of WMT}.

\bibitem[{Post(2018)}]{matt2018clarity}
Matt Post. 2018.
\newblock A call for clarity in reporting {BLEU} scores.
\newblock In \emph{Proceedings of the Third Conference on Machine Translation:
  Research Papers}. Association for Computational Linguistics.

\bibitem[{Sennrich et~al.(2016)Sennrich, Haddow, and
  Birch}]{sennrich2016neural}
Rico Sennrich, Barry Haddow, and Alexandra Birch. 2016.
\newblock Neural machine translation of rare words with subword units.
\newblock In \emph{Proceedings of the 54th Annual Meeting of the Association
  for Computational Linguistics}, pages 1715--1725.

\bibitem[{Vaswani et~al.(2017)Vaswani, Shazeer, Parmar, Uszkoreit, Jones,
  Gomez, Łukasz Kaiser, and Polosukhin}]{vaswani2017attention}
Ashish Vaswani, Noam Shazeer, Niki Parmar, Jakob Uszkoreit, Llion Jones, Aidan
  Gomez, Łukasz Kaiser, and Illia Polosukhin. 2017.
\newblock Attention is all you need.
\newblock In \emph{Proceedings of the 31st Conference on Neural Information
  Processing Systems}.

\end{thebibliography}
\bibliographystyle{acl_natbib}

\end{document}